\documentclass[sigconf]{acmart}

\AtBeginDocument{%
  \providecommand\BibTeX{{%
    \normalfont B\kern-0.5em{\scshape i\kern-0.25em b}\kern-0.8em\TeX}}}

\copyrightyear{2023}
\acmYear{2023}
\setcopyright{rightsretained}
\acmConference[MM '23]{Proceedings of the 31st ACM International Conference on Multimedia}{October 29-November 3, 2023}{Ottawa, ON, Canada}
\acmBooktitle{Proceedings of the 31st ACM International Conference on Multimedia (MM '23), October 29-November 3, 2023, Ottawa, ON, Canada}
\acmDOI{10.1145/3581783.3612365}
\acmISBN{979-8-4007-0108-5/23/10}

\usepackage{amsthm,amsmath,amssymb}
\usepackage{mathrsfs}
\usepackage{booktabs}
\usepackage{multirow}
\usepackage{multicol}
\usepackage{xcolor}
\usepackage{color}
\usepackage{hyperref}
\usepackage{float}
\usepackage{subfig}

\newcommand{\tabincell}[2]{\begin{tabular}{@{}#1@{}}#2\end{tabular}}

\begin{document}


\title{MAE-DFER: Efficient Masked Autoencoder for Self-supervised Dynamic Facial Expression Recognition}

\author{Licai Sun}
\email{sunlicai2019@ia.ac.cn}
\affiliation{%
  \institution{School of Artificial Intelligence, University of Chinese Academy of Sciences}
  \institution{Institute of Automation, Chinese Academy of Sciences}
  \city{Beijing}
  \country{China}
}

\author{Zheng Lian}
\email{lianzheng2016@ia.ac.cn}
\affiliation{%
  \institution{Institute of Automation, Chinese Academy of Sciences}
  \city{Beijing}
  \country{China}
}

\author{Bin Liu}
\email{liubin@nlpr.ia.ac.cn}
\affiliation{%
  \institution{Institute of Automation, Chinese Academy of Sciences}
  \institution{School of Artificial Intelligence, University of Chinese Academy of Sciences}
  \city{Beijing}
  \country{China}
}

\author{Jianhua Tao}
\email{jhtao@tsinghua.edu.cn}
\affiliation{%
  \institution{Department of Automation, Tsinghua University}
  \institution{Beijing National Research Center for Information Science and Technology, Tsinghua University}
  \city{Beijing}
  \country{China}
}


\begin{abstract}
Dynamic facial expression recognition (DFER) is essential to the development of intelligent and empathetic machines. Prior efforts in this field mainly fall into supervised learning paradigm, which is severely restricted by the limited labeled data in existing datasets. Inspired by recent unprecedented success of masked autoencoders (e.g., VideoMAE), this paper proposes MAE-DFER, a novel self-supervised method which leverages large-scale self-supervised pre-training on abundant unlabeled data to largely advance the development of DFER. Since the vanilla Vision Transformer (ViT) employed in VideoMAE requires substantial computation during fine-tuning, MAE-DFER develops an efficient local-global interaction Transformer (LGI-Former) as the encoder. Moreover, in addition to the standalone appearance content reconstruction in VideoMAE, MAE-DFER also introduces explicit temporal facial motion modeling to encourage LGI-Former to excavate both static appearance and dynamic motion information. Extensive experiments on six datasets show that MAE-DFER consistently outperforms state-of-the-art supervised methods by significant margins (e.g., +6.30\% UAR on DFEW and +8.34\% UAR on MAFW), verifying that it can learn powerful dynamic facial representations via large-scale self-supervised pre-training. Besides, it has comparable or even better performance than VideoMAE, while largely reducing the computational cost (about 38\% FLOPs). We believe MAE-DFER has paved a new way for the advancement of DFER and can inspire more relevant research in this field and even other related tasks. Codes and models are publicly available at \textcolor[rgb]{0.93,0.0,0.47}{\url{https://github.com/sunlicai/MAE-DFER}}.

\end{abstract}




\maketitle

\section{Introduction}
Facial expressions, as an important aspect of nonverbal communication, play a significant role in interpersonal interactions \cite{darwin1998expression}. 
In the past two decades, automatic facial expression recognition (FER) has drawn widespread attention due to its crucial role in developing intelligent and empathetic machines that can interact with humans in a natural and intuitive way \cite{pantic2000automatic, picard2000affective, de2023social}. 
FER also has a wide spectrum of practical applications in areas such as healthcare \cite{bisogni2022impact}, education \cite{whitehill2014faces}, and entertainment \cite{tao2005affective}.
According to the input data type, FER can be divided into two categories, i.e., static FER (SFER) and dynamic FER (DFER) \cite{li2020deep}. SFER takes static facial images as input, while DFER aims to recognize expressions in dynamic image sequences or videos. 
Since SFER overlooks the critical temporal information for the interpretation of facial expressions, 
this paper mainly focuses on DFER.

\begin{figure}[t]
	\centering
	\includegraphics[width=\linewidth]{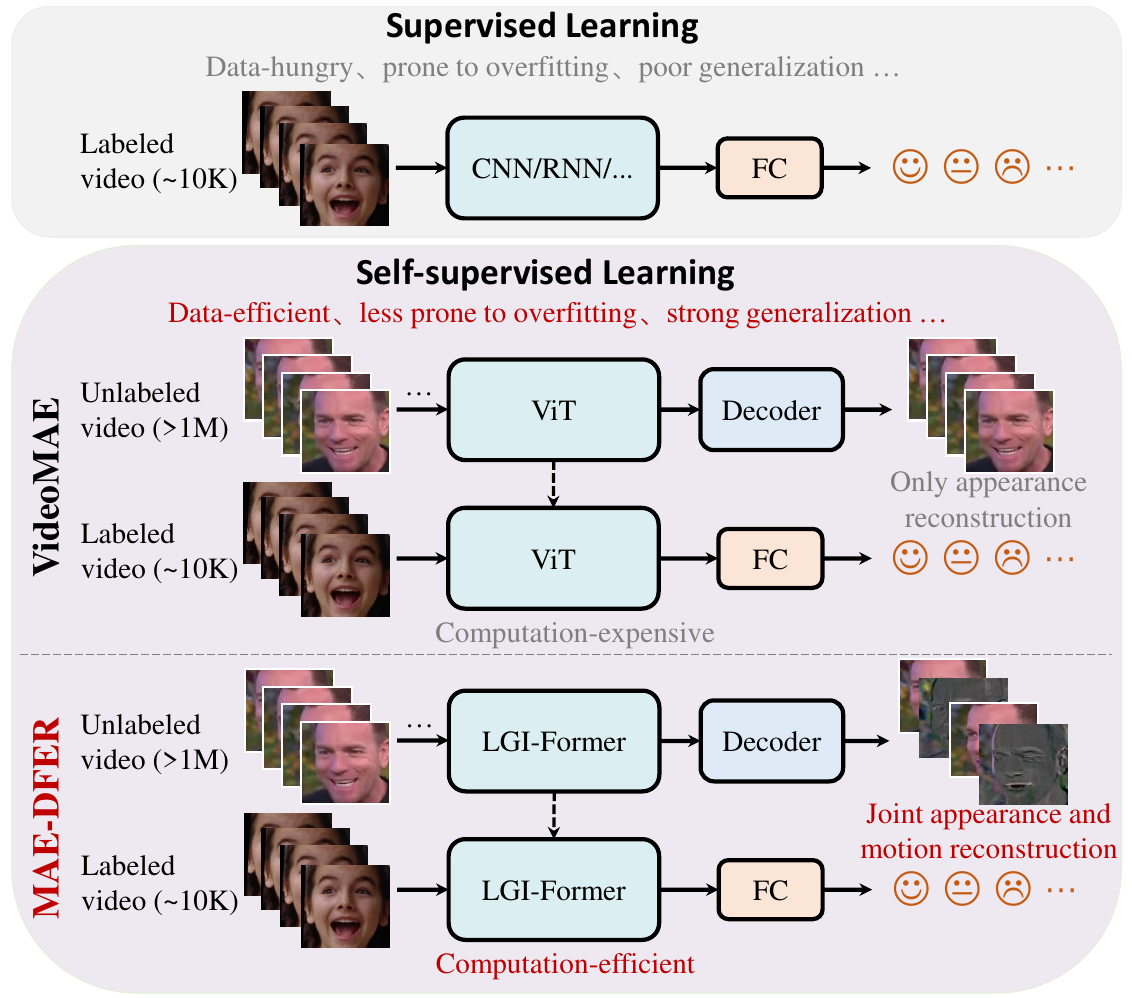}
	\caption{An overview of the proposed MAE-DFER.}
	\label{fig_overview}
\end{figure}

DFER is dominated by the supervised learning paradigm. 
Researchers have developed various deep neural networks for this task, including 2D/3D convolutional neural networks (CNN) \cite{fan2016video, jiang2020dfew, kossaifi2020factorized}, recurrent neural networks (RNN) \cite{ebrahimi2015recurrent, sun2020multi, wang2022dpcnet}, and more advanced Transformer-based architectures \cite{zhao2021former, ma2022spatio, wang2022emotional, liu2023expression, li2023intensity}. 
Although supervised methods have achieved remarkable success, the limited training samples in existing DFER datasets (typically around 10K, which is much smaller than those in other research areas such as general image/video classification and face recognition, see details in Table \ref{tab_dataset}) severely restrict their further advancement (e.g., training large video Transformers). A straightforward idea to address this issue is to increase the dataset scale. However, collecting and annotating large-scale high-quality DFER datasets is pretty time-consuming and labor-intensive, which is mainly due to the sparsity of dynamic facial expressions in videos and the ambiguity and subjectivity in facial expression perception \cite{li2020deep, jiang2020dfew, wang2022ferv39k}. Considering that there are massive unlabeled facial videos on the Internet, a natural question arises in the mind: \textit{can we exploit them to fully unleash the power of deep neural networks for better DFER?}

The recent progress of self-supervised learning in many deep learning fields \cite{devlin2018bert, he2022masked, baevski2020wav2vec} indicates that there is a positive answer. Notably, masked autoencoder (MAE) \cite{he2022masked} in computer vision develops an asymmetric encoder-decoder architecture for masked image modeling. It successfully pre-trains the vanilla Vision Transformer (ViT) \cite{dosovitskiy2020image} in an end-to-end manner and outperforms the supervised baselines in many vision tasks. Subsequently, VideoMAE \cite{tong2022videomae} extends MAE to the video domain and also achieves impressive results on lots of general video datasets. Motivated by this line of research, we present MAE-DFER (Fig. \ref{fig_overview}), a novel self-supervised method based on VideoMAE which leverages large-scale self-supervised pre-training on abundant unlabeled facial video data to promote the advancement of DFER. Although VideoMAE has made remarkable success in self-supervised video pre-training, we notice that it still has two main drawbacks: 1) The vanilla ViT encoder employed in VideoMAE requires substantial computation during fine-tuning due to the quadratic scaling cost of global space-time self-attention, which is unaffordable in many resource-constrained scenarios. 2) It only reconstructs video appearance contents during pre-training, thus might be insufficient to model temporal facial motion information which is also crucial to DFER.

To tackle the above issues in VideoMAE, our MAE-DFER presents two core designs accordingly. 
For the \textit{first} issue, MAE-DFER develops an \textit{efficient} local-global interaction Transformer (LGI-Former) as the encoder. Different from the global space-time self-attention in ViT, LGI-Former first constrains self-attention in local spatiotemporal regions and then utilizes a small set of learnable \textit{representative tokens} to enable efficient local-global information exchange. Concretely, it decomposes the global space-time self-attention into three stages: local intra-region self-attention, global inter-region self-attention, and local-global interaction. 
In this way, LGI-Former can efficiently propagate global information to local regions and avoid the expensive computation of global space-time attention.
For the \textit{second} issue, MAE-DFER introduces \textit{joint} masked appearance and motion modeling to encourage the model to capture both static facial appearance and dynamic motion information. Specifically, in addition to the original appearance content reconstruction branch, it simply utilizes the frame difference signal as another reconstruction target for explicit temporal facial motion modeling. 
To verify the effectiveness of MAE-DFER, we perform large-scale self-supervised pre-training on the VoxCeleb2 dataset \cite{chung2018voxceleb2}, which has more than 1M unlabeled facial video clips collected from YouTube. 
Then we fine-tune the pre-trained model on six DFER datasets, including three relatively large in-the-wild datasets (DFEW \cite{jiang2020dfew}, FERV39k \cite{wang2022ferv39k}, and MAFW \cite{liu2022mafw}) and three small lab-controlled datasets (CREMA-D \cite{cao2014crema}, RAVDESS \cite{livingstone2018ryerson}, and eNTERFACE05 \cite{martin2006enterface}). The results show that MAE-DFER significantly outperforms the state-of-the-art supervised methods, indicating that it is capable of learning strong and useful dynamic facial representations for DFER. Moreover, compared with VideoMAE, MAE-DFER largely reduces \textbf{$\sim$38\%} FLOPs while having comparable or even better performance. 
The main contributions of this paper are summarized as follows:
\begin{itemize}
\item We present a novel self-supervised method, MAE-DFER, as an early attempt to leverage large-scale self-supervised pre-training on abundant unlabeled facial video data to advance the development of DFER. 
\item MAE-DFER improves VideoMAE by developing an efficient LGI-Former as the encoder and introducing joint masked appearance and motion modeling. With these two core designs, MAE-DFER largely reduces the computational cost while having comparable or even better performance.
\item Extensive experiments on six DFER datasets show that our MAE-DFER consistently outperforms the previous best supervised methods by significant margins (\textbf{+5$\sim$8\%} UAR on three in-the-wild datasets and \textbf{+7$\sim$12\%} WAR on three lab-controlled datasets), which demonstrates that it can learn powerful dynamic facial representations for DFER via large-scale self-supervised pre-training. 

\end{itemize}

\section{Related Work}

\subsection{Dynamic Facial Expression Recognition}
The early studies on DFER primarily focus on designing various local descriptors and only several very small lab-controlled datasets are available for evaluation. 
With the emergence of deep learning and the proliferation of relatively larger datasets, the research paradigm has undergone a transformative shift towards training deep neural networks in an end-to-end fashion. 
In general, there are three trends. 
The first trend directly utilizes 3D CNNs (such as C3D \cite{tran2015learning}, 3D ResNet \cite{hara2018can}, R(2+1)D \cite{tran2018closer}, and P3D \cite{qiu2017learning}) to extract joint spatiotemporal features  from raw facial videos \cite{fan2016video, jiang2020dfew, kossaifi2020factorized, wang2022ferv39k, liu2022mafw, wang2023rethinking}. 
The second trend uses the combination of 2D CNN (e.g., VGG \cite{simonyan2014very} and ResNet \cite{he2016deep}) and RNN (e.g., LSTM \cite{hochreiter1997long} and GRU \cite{chung2014empirical}) \cite{ebrahimi2015recurrent, kollias2020exploiting, jiang2020dfew, sun2020multi, wang2022ferv39k, liu2022mafw, wang2022dpcnet}. 
Recently, with the rise of Transformer \cite{vaswani2017attention}, several studies exploit its global dependency modeling ability to augment CNN/RNN for better performance, which forms the third trend \cite{zhao2021former, ma2022spatio, liu2023expression, li2022nr, liu2022mafw, li2023intensity}. 
For instance, Former-DFER \cite{zhao2021former} employs a Transformer-enhanced ResNet-18 for spatial feature extraction and another Transformer for temporal information aggregation. 
STT \cite{ma2022spatio} improves Former-DFER by introducing factorized spatial and temporal attention for joint spatiotemporal feature learning. 
IAL \cite{li2023intensity} further introduces the global convolution-attention block and intensity-ware loss to deal with expressions with different intensities.
However, all the above methods fall into the supervised learning paradigm, which is thus restricted by the limited training samples in existing DFER datasets. Unlike them, this paper proposes a self-supervised method that can learn powerful representations from massive unlabeled facial video data and achieve significant improvement over them.

\subsection{Masked Autoencoders}
Masked autoencoders (MAEs), as the representative of generative self-supervised learning, have recently achieved unprecedented success in many deep learning fields \cite{zhang2022survey}. They are mainly inspired by the progress of masked language modeling (e.g., BERT \cite{devlin2018bert} and GPT \cite{radford2018improving}) in natural language processing and typically adopt a mask-then-predict strategy to pre-train the vanilla ViT. Notably, iGPT \cite{chen2020generative} follows GPT to auto-regressively predict pixels and makes the first successful attempt. BEiT \cite{bao2022beit} follows BERT and adopts a two-stage training pipeline, i.e., first utilizing an off-the-shelf tokenizer to generate discrete visual tokens and then performing masked-then-predict training. MAE \cite{he2022masked} improves BEiT by designing an asymmetric encoder-decoder architecture to enable efficient end-to-end pre-training. After that, many studies adopt the architecture of MAE to perform self-supervised pre-training on various tasks. For instance, VideoMAE \cite{tong2022videomae} and its concurrent work MAE-ST \cite{feichtenhofer2022masked} extends MAE to the video domain and achieve impressive results on lots of video benchmarks. 
Our proposed MAE-DFER is inspired by VideoMAE and it develops two core designs to facilitate effective and efficient representation learning for DFER.

\section{Method}

\begin{figure*}[t]
	\centering
	\includegraphics[height=0.4\linewidth,width=0.9\linewidth]{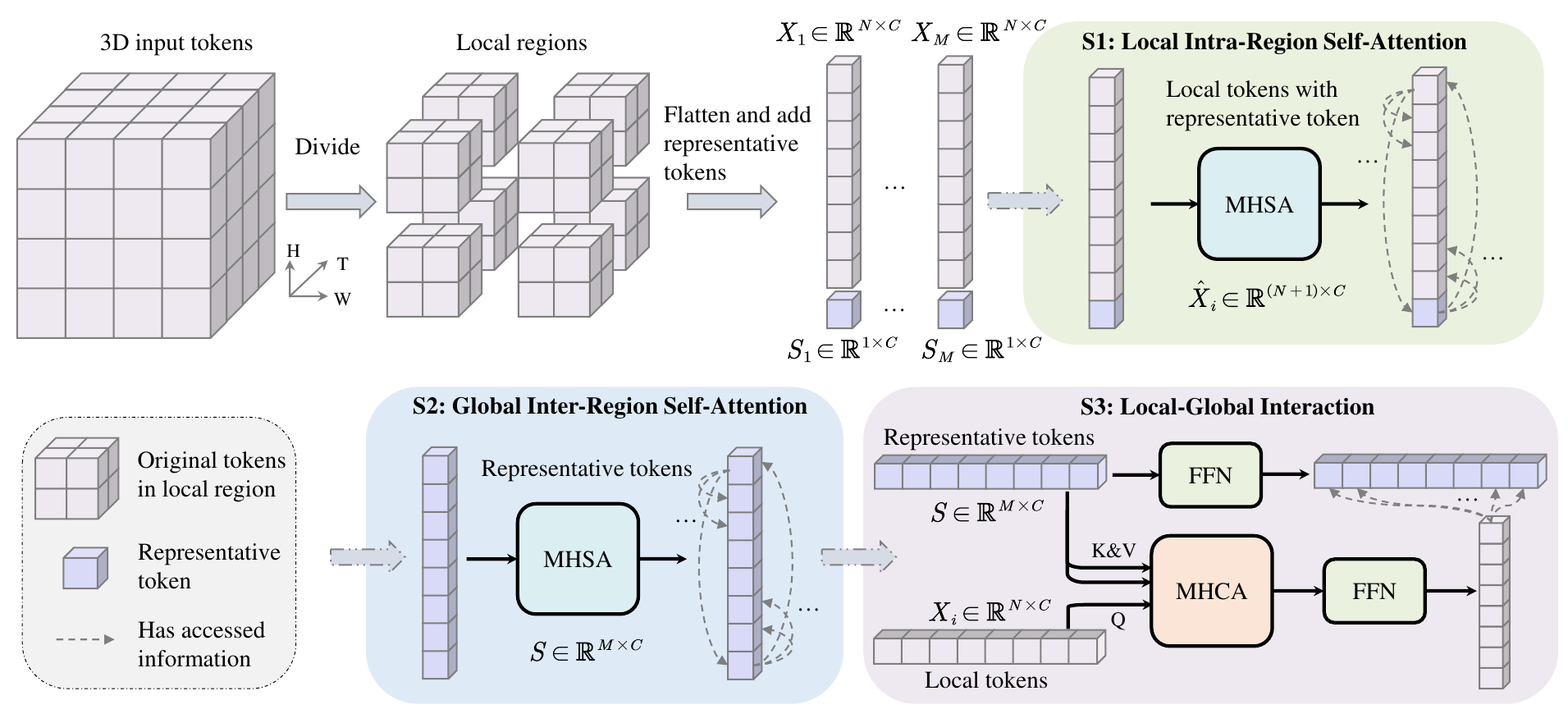}
	\caption{The illustration of LGI-Former. For simplicity, we only present the information flow in one block, which mainly consists of three stages: 1) local intra-region self-attention, 2) global inter-region self-attention, and 3) local-global interaction.}
	\label{fig_encoder}
\end{figure*}

\subsection{Revisiting VideoMAE}
\label{sec_videomae}
VideoMAE \cite{tong2022videomae} is a simple extension of MAE \cite{he2022masked} in the video domain. It basically follows the asymmetric encoder-decoder architecture of MAE for self-supervised video pre-training.
The main difference is that a much higher masking ratio (i.e., 90\% vs. 75\%) and tube masking strategy (instead of random masking) are adopted, considering that large temporal redundancy and high temporal correlation in videos \cite{tong2022videomae}.
In specific, VideoMAE mainly consists of four modules: cube embedding, tube masking, a high-capacity encoder $\Phi_e$ (i.e., the vanilla ViT), and a lightweight decoder $\Phi_d$.
Given a raw video $\mathbf{V} \in \mathbb{R}^{T \times H \times W \times 3}$, VideoMAE first utilizes cube embedding with a cube size of $2 \times 16 \times 16$ to transform $\mathbf{V}$ into a sequence of tokens $\mathbf{X} \in \mathbb{R}^{K \times C}$, where $K=\frac{T}{2} \cdot \frac{H}{16} \cdot \frac{W}{16}$ and $C$ is the channel size. 
Then the tube masking module generates a mask $\mathbf{M} \in \{0, 1\}^K$ with a masking ratio of $\rho=90\%$ and the high-capacity encoder $\Phi_e$ only takes the unmasked tokens $\mathbf{X} \odot \mathbf{M} \in \mathbb{R}^{L \times C}$ ($L = (1-\rho)K$) as input and simply process them with global space-time self-attention. Subsequently, the lightweight decoder $\Phi_d$ combines the encoded visible tokens with the learnable mask tokens (with a size of $\rho K$) to reconstruct the raw video data. 
Finally, the mean square error between the original and reconstructed video in the masked positions are calculated to optimize the whole model. The above process can be generally formulated as follows:
\begin{equation}
\mathcal{L}_{\textrm{VideoMAE}} = \textrm{MSE}(\Phi_d(\Phi_e(\mathbf{X} \odot \mathbf{M})), \mathbf{V} \odot \Psi(1-\mathbf{M}))
\label{eq_videomae_loss}
\end{equation}
where $\Psi$ is a function used to obtain masked positions in the pixel space.
In downstream tasks, the lightweight decoder $\Phi_d$ is discarded and only the high-capacity ViT encoder $\Phi_e$ will be fine-tuned.

\subsection{MAE-DFER: Overview}
\label{sec_overview}
Although VideoMAE has made great success in self-supervised video pre-training, it still faces two major challenges. First, it only focuses on reconstructing raw appearance contents in the video, which thus lacks explicit temporal motion modeling and might not be sufficient to model temporal facial motion information. Second, although it enjoys high efficiency during \textit{pre-training} through an asymmetric encoder-decoder architecture (i.e., dropping a large proportion of masked tokens to save computation), the computational cost of global space-time self-attention in the vanilla ViT is still extremely expensive during downstream \textit{fine-tuning} since it cannot drop input tokens at this stage. To tackle these issues, as shown in Fig. \ref{fig_overview}, we propose MAE-DFER, a new self-supervised framework for DFER. For the first issue, MAE-DFER introduces joint masked appearance and motion modeling to encourage the model to excavate both static appearance and dynamic motion information (Section \ref{sec_joint_modeling}). For the second issue, it employs a novel Local-Global Interaction Transformer (LGI-Former) as the encoder to largely reduce the computational cost of ViT during downstream fine-tuning (Section \ref{sec_lgi_former}).

\subsection{MAE-DFER: Joint Masked Appearance and Motion Modeling}
\label{sec_joint_modeling}
Temporal motion information matters for DFER (e.g., the gradual appearance and disappearance of a smile may convey totally different emotions).
To explicitly incorporate this information in self-supervised pre-training, our MAE-DFER adds an additional temporal motion reconstruction branch in parallel with the original appearance reconstruction branch in VideoMAE to achieve \textit{joint} facial appearance and motion structure learning. Specifically, we simply calculate the frame difference signal as the temporal motion target given that its computation is very cheap and it has shown effectiveness in video action recognition \cite{wang2016temporal, wang2021tdn, song2022takes}. To ensure that the computational cost during pre-training similar to VideoMAE, we share the decoder backbone for appearance and motion branches and only use two different linear heads to predict their targets.
Besides, the decoder only outputs appearance predictions in the odd frames and motion predictions in the remaining even frames. Finally, the total loss is the weighted sum of mean square errors in two branches:
\begin{equation}
\begin{split}
\mathcal{L}_{\textrm{MAE-DFER}} = \lambda &\cdot \textrm{MSE}(\Phi_d(\Phi_e(\mathbf{X} \odot \mathbf{M})), \mathbf{V}_a \odot \Psi(1-\mathbf{M})) + \\
 (1-\lambda) &\cdot \textrm{MSE}(\Phi_d(\Phi_e(\mathbf{X} \odot \mathbf{M})), \mathbf{V}_m \odot \Psi(1-\mathbf{M}))
\end{split}
\label{eq_mae_dfer_loss}
\end{equation}
where $\mathbf{V}_a = \mathbf{V}[0\colon T \colon 2]$ is the appearance target, $\mathbf{V}_m = \mathbf{V}[1 \colon T \colon 2] - \mathbf{V}[0 \colon T \colon 2]$ is the motion target, $\lambda$ is a hyperparameter to balance the contribution of two branches and we empirically set it to 0.5.

\subsection{MAE-DFER: Efficient LGI-Former}
\label{sec_lgi_former}
The architecture of LGI-Former is illustrated in Fig. \ref{fig_encoder}.
Unlike the global space-time self-attention adopted in the vanilla ViT, LGI-Former constrains self-attention in local spatiotemporal regions to save computation. However, simply stacking multiple local self-attention layers does not permit inter-region information exchange.
Inspired by \cite{fang2022msg} and \cite{sun2023efficient}, the core idea of LGI-Former is to introduce a small set of \textit{representative tokens} to local regions.
On the one hand, these tokens take charge of summarizing critical information in local regions. On the other hand, they allow for long-range dependencies modeling between different regions and enable efficient local-global information exchange. 
Thanks to the introduction of representative tokens, the expensive global space-time self-attention can be decomposed into three stages with much cheaper computation: 1) local intra-region self-attention, 2) global inter-region self-attention, and 3) local-global interaction.
In the following, for simplicity, we only describe the above three stages during fine-tuning. The process during pre-training is similar as MAE-DFER follows VideoMAE to adopt the tube masking strategy and applies the same masking ratio to each local region to ensure that all regions have an equal number of visible tokens.
 
\textbf{Local Intra-Region Self-Attention.}
For convenience, we first reshape the input sequence $\mathbf{X} \in \mathbb{R}^{K \times C}$ (after cube embedding) to 3D tokens $\mathbf{X} \in \mathbb{R}^{\frac{T}{2} \times \frac{H}{16} \times \frac{W}{16} \times C}$ and divide it into non-overlapped local spatiotemporal regions with an equal size of $t \times h \times w$ as shown in Fig. \ref{fig_encoder}.
In each region, apart from the original tokens, we also add a learnable representative token. 
The local intra-region self-attention then operates on their concatenation to simultaneously promote fine-grained local feature learning and enable local information aggregation into the representative token. 
Assume that the original local tokens and the associated representative token in the $i$th region is $\mathbf{X}_i \in \mathbb{R}^{N \times C}$ and $\mathbf{S}_i \in \mathbb{R}^{1 \times C}$ respectively ($N=thw$, $i \in \{1,2, ..., M\}$, and $M=\frac{K}{N}$ is the number of representative tokens), the formulation of local intra-region self-attention is given as follows:
\begin{align}
\mathbf{\hat{X}}_i &= \textrm{Concat}(\mathbf{S}_i, \mathbf{X}_i) \\
\mathbf{\hat{X}}_i &= \textrm{MHSA}(\textrm{LN}(\mathbf{\hat{X}}_i)) + \mathbf{\hat{X}}_i 
\end{align}
where $\mathbf{\hat{X}}_i \in  \mathbb{R}^{(N+1) \times C}$, $\textrm{MHSA}$ is the multi-head self-attention in the vanilla ViT, and $\textrm{LN}$ stands for layer normalization. In particular, the calculation of $\textrm{MHSA}$ is formulated as follows:
\begin{align}
\textrm{MHSA}(\mathbf{X}) &= \textrm{Concat}(\textrm{head}_{1}, ..., \textrm{head}_{h})\mathbf{W}^O \\
\textrm{head}_{j} &= \textrm{Attention}(\mathbf{X} \mathbf{W}^{Q}_j, \mathbf{X} \mathbf{W}^{K}_j, \mathbf{X} \mathbf{W}^{V}_j) \\
\textrm{Attention}(\mathbf{Q}, \mathbf{K}, \mathbf{V}) &= \textrm{softmax}(\frac{\mathbf{Q} \mathbf{K}^\top}{\sqrt{d}})\mathbf{V}
\end{align}
where $\mathbf{W}^{*}_j \in \mathbb{R}^{C\times d}$ ($* \in \{Q,K,V \}$), $\mathbf{W}^{O} \in \mathbb{R}^{C\times C}$, $h$ is the number of attention heads, $d=\frac{C}{h}$ is the feature dimension of each head.

\textbf{Global Inter-Region Self-Attention.}
After local intra-region self-attention, the representative token has extracted crucial information in each local region and can \textit{represent} the original tokens to perform information exchange between different regions. 
Since the number of representative tokens is typically small (e.g., 8), the computational cost for inter-region communication can be negligible.
Thus, we first aggregate all representative tokens and then simply utilize global inter-region self-attention on them to propagate information between different regions, i.e., 
\begin{align}
\mathbf{S} &= \textrm{Concat}(\mathbf{S}_1, ..., \mathbf{S}_M) \\
\mathbf{S} &= \textrm{MHSA}(\textrm{LN}(\mathbf{S})) + \mathbf{S}
\end{align}
where $\mathbf{S} \in  \mathbb{R}^{M \times C}$ is the aggregated representative tokens. 

\textbf{Local-Global Interaction.}
After information propagation via global inter-region self-attention, the representative token in each local region has been consolidated by useful information from other regions, thus having a global view of the whole input tokens.
To enable the original tokens in each local region to access the global information, we further employ cross-attention between local tokens and representative tokens to achieve that goal:
\begin{align}\label{eq_lobal_global}
\mathbf{X}_i &= \textrm{MHCA}(\textrm{LN}(\mathbf{X}_i), \textrm{LN}(\mathbf{S})) + \mathbf{X}_i \\
\mathbf{X}_i &= \textrm{FFN}(\textrm{LN}(\mathbf{X}_i)) + \mathbf{X}_i \\
\mathbf{S} &= \textrm{FFN}(\textrm{LN}(\mathbf{S})) + \mathbf{S}
\end{align}
where $\textrm{MHCA}$ is multi-head cross-attention and $\textrm{FFN}$ denotes feed-forward network. Specifically, $\textrm{MHCA}$ has the similar implementation with $\textrm{MHSA}$ except that its query and key/value come from different inputs, i.e., 
\begin{align}
\textrm{MHCA}(\mathbf{X}, \mathbf{Y}) &= \textrm{Concat}(\textrm{head}_{1}, ..., \textrm{head}_{h})\mathbf{W}^O \\
\textrm{head}_{h} &= \textrm{Attention}(\mathbf{X} \mathbf{W}^{Q}_j, \mathbf{Y} \mathbf{W}^{K}_j, \mathbf{Y} \mathbf{W}^{V}_j)
\end{align}

\textbf{Complexity Analysis.}
We suppose that the flattened input is $\mathbf{X} \in \mathbb{R}^{K \times C}$, where $K=MN$ is the  number of total input tokens, $M$ is the number of local regions and $N$ is the number of original tokens in each region. 
Since self-attention scales quadratically with the sequence length, the complexity of local intra-region self-attention is $O(M(N+1)^2) \approx O(MN^2) = O(\frac{K^2}{M})$. Similarly, the complexity of global inter-region self-attention is $O(M^2) = O(\frac{K^2}{N^2})$. Moreover, local-global interaction has a complexity of $O(MNM)=O(\frac{K^2}{N})$. Putting them together, the complexity of an LGI-Former block is $O((\frac{1}{M} + \frac{1}{N^2} + \frac{1}{N}) K^2$), while a standard Transformer block in the vanilla ViT has a complexity of $O(K^2)$. In practice, $M \ll K$ and $N \ll K$, thus the computational cost of LGI-Tranformer is largely reduced compared with the vanilla ViT.

\section{Results}

\subsection{Datasets}
\textbf{Pre-training Dataset.} We perform self-supervised pre-training on VoxCeleb2 
 \cite{chung2018voxceleb2}. It has over 1 million video clips of more than 6,000 celebrities, extracted from around 150,000 interview videos on YouTube. It is divided into a development set and a test set. We only use the \textit{development} set for pre-training, which contains 1,092,009 video clips from 145,569 videos. 

\textbf{DFER Datasets.} We conduct experiments on 6 datasets, including 3 large \textit{in-the-wild} datasets (i.e., DFEW \cite{jiang2020dfew}, FERV39k \cite{wang2022ferv39k}, and MAFW \cite{liu2022mafw}) and 3 small \textit{lab-controlled} datasets (i.e., CREMA-D \cite{cao2014crema}, RAVDESS \cite{livingstone2018ryerson}, and eNTERFACE05 \cite{martin2006enterface}). Their basic information is summarized in Table \ref{tab_dataset}. 
Detailed introductions can be found in Appendix \ref{sec_appendix_dataset}.
Following previous studies \cite{jiang2020dfew, zhao2021former, liu2022mafw, wang2022ferv39k}, we report both unweighted average recall (UAR, i.e., the mean class accuracy) and weighted average recall (WAR, i.e., the overall accuracy). For those datasets using cross-validation, we combine the predictions and labels from all folds to calculate the final UAR and WAR.

\begin{table}[]
\caption{Basic information of six DFER datasets used in this paper. 
CV: cross-validation. $^\dag$: subject-independent setting.}
\label{tab_dataset}
\centering
\resizebox{\linewidth}{!}{
\begin{tabular}{lcrcccc}
\toprule
Name & Wild & \#Videos & \#Classes & Evaluation \\
\midrule
DFEW \cite{jiang2020dfew}       &  $\checkmark$ &  11,697  &  7   & Default 5-fold CV\\
FERV39k \cite{wang2022ferv39k}  &  $\checkmark$ & 38,935  &  7   & Default train \& test   \\
MAFW  \cite{liu2022mafw}        &  $\checkmark$ & 9,172   &  11  & Default 5-fold CV \\
CREMA-D  \cite{cao2014crema}    &  $\times$ &  7,442  &  6   & 5-fold CV $^\dag$ \\
RAVDESS  \cite{livingstone2018ryerson}  & $\times$  &    1,440  &  8 & 6-fold CV $^\dag$ \\
eNTERFACE05 \cite{martin2006enterface}  & $\times$  &    1,287  &  6 & 5-fold CV $^\dag$ \\
\bottomrule
\end{tabular}
}
\end{table}

\subsection{Implementation Details}
\textbf{MAE-DFER.} For the high-capacity encoder, we adopt the LGI-Former which has 16 blocks and a hidden size of 512. The total number of parameters is 84.9M, which is similar to that (86.2M) of ViT base model. The local region size is set to $2\times5\times10$ by default. 
For the lightweight decoder, we follow VideoMAE to adopt four standard Transformer blocks with a hidden size of 384.

\textbf{Pre-training.} The original videos provided in VoxCeleb2 have a resolution of $224\times224$. Given that the speaker's face generally does not fill the entire frame, we only used a $160\times160$ patch located in the upper center of each video frame to remove the irrelevant background information.
During pre-training, we extract 16 frames from each video clip using a temporal stride of 4. This results in $8\times10\times10$ input tokens after cube embedding, when using a cube size of $2\times16\times16$. 
Regarding hyperparameters, we mainly follow VideoMAE. Specifically, we use an AdamW optimizer with $\beta_1=0.9$ and $\beta_2=0.95$, an overall batch size of 128, a base learning rate of $3e-4$, and a weight decay of 0.05. We linearly scale the base learning rate according to the overall batch size, using the formula: $\textrm{lr} = \textrm{base learning rate} \times \frac{\textrm{batch size}}{256}$. In addition, we use a cosine decay learning rate scheduler. By default, we pre-train the model for 50 epochs, with 5 warmup epochs. When using 4 Nvidia Tesla V100 GPUs, the pre-training takes about 3-4 days.

\textbf{Fine-tuning.} 
Same as pre-training, the input clip size is $16\times160\times160$ and the temporal stride is 4 for most datasets (except 1 for FERV39k). To optimize the model, we use an AdamW optimizer with $\beta_1=0.9$ and $\beta_2=0.999$, with a base learning rate of $1e-3$ and an overall batch size of 96. The other hyperparameters remain the same as in pre-training, and more details can be found in \cite{tong2022videomae}. We fine-tune the pre-trained model for 100 epochs, with 5 warmup epochs. During inference, we sample two clips uniformly along the temporal axis for each video and then calculate the average score as the final prediction.

\subsection{Ablation Studies}
In this part, we conduct ablation experiments on DFEW and FERV39k to demonstrate the effects of several key factors in MAE-DFER. For simplicity, on DFEW, we only report results of fold 1 (fd1).

\textbf{Pre-training Epochs.}
As shown in Table \ref{tab_ablation_pretraining_epochs}, we observe that longer pre-training is generally beneficial and the performance saturation occurs at around 50 epochs. Besides, we also find that the performance of training from scratch (i.e., \#Epochs=0) is very poor (nearly random guessing). This is largely attributed to the limited training samples in current DFER datasets since large vision Transformers are data-hungry and training them typically requires more than million-level labeled data \cite{dosovitskiy2020image, tong2022videomae}. This result also demonstrates the significance and superiority of large-scale self-supervised pre-training over traditional supervised learning. 

\begin{table}[]
\caption{Ablation study on the pre-training epochs.}
\label{tab_ablation_pretraining_epochs}
\centering
\scalebox{0.85}{
\begin{tabular}{ccccccc}
\toprule
\multirow{2}{*}{Dataset} & \multirow{2}{*}{Metric} & \multicolumn{5}{c}{Pre-training Epochs}  \\
\cmidrule(lr){3-7}
    &     & 0    & 10    & 30    & 50 & 70  \\
\midrule
\multirow{2}{*}{DFEW} & UAR  & 15.08 & 58.05 & 60.46 & \textbf{62.59} & 61.93 \\
                      & WAR  & 23.84 & 71.29 & 72.73 & \textbf{74.88} & 74.58 \\
\midrule                  
\multirow{2}{*}{FERV39k} & UAR  & 18.09 & 39.98 & 42.04 & 43.12 & \textbf{43.15} \\
                         & WAR  & 28.33 & 50.21 & 51.62 & \textbf{52.07} & 52.01 \\
\bottomrule
\end{tabular}
}
\end{table}


\textbf{Comparison of Different Model Architectures.}
We then investigate the effect of three key modules in LGI-Former by evaluating the performance of the following variants: 1) only local intra-region self-attention (i.e., no global inter-region self-attention and local-global interaction), 2) no local-global interaction, 3) no global inter-region self-attention, and 4) using global space-time self-attention instead (i.e., ViT). The results are presented in Table \ref{tab_ablation_model_arch}. We have the following observations: 1) The first variant has the worst performance, which is as expected since only utilizing local intra-region self-attention does not allow local tokens to access global information. 2) Either global inter-region self-attention or local-global interaction contributes to better performance, demonstrating the effectiveness of these two modules in local-global information propagation. Besides, the latter is generally more effective than the former but at the cost of more computation. It also should be noted that global inter-region self-attention only introduces negligible computation ($\sim$0.1G FLOPs) thanks to the small number (i.e., 8) of representative tokens. 3) When combing the global inter-region self-attention with local-global interaction, LGI-Former achieves the best results. Besides, compared with the last variant which uses global space-time self-attention (i.e., ViT), we only observe slight performance drop (<0.6\%) but large computation reduction ($\sim$38\% FLOPs), thus demonstrating the efficiency of LGI-Former.

\begin{table}[]
\caption{Ablation study on the model architecture. Intra: local intra-region self-attention. Inter: global inter-region self-attention. LGI: local-global interaction.}
\label{tab_ablation_model_arch}
\centering
\resizebox{\linewidth}{!}{
\begin{tabular}{lccccccc}
\toprule
Dataset & Intra &  Inter &  LGI &  \tabincell{c}{\#Params (M)}  & \tabincell{c}{FLOPs (G)}   & UAR    & WAR  \\
\midrule
\multirow{4}{*}{DFEW} 
& \checkmark   & $\times$     & $\times$       &  51.2    & 42.7    & 59.66  & 72.00     \\
& \checkmark   & \checkmark   & $\times$       &  68.0    & 42.8    & 60.43  & 73.69     \\
& \checkmark   & $\times$     & \checkmark     &  68.1    & 49.6    & 60.98  & 74.58     \\
& \checkmark   & \checkmark   & \checkmark     &  84.9    & 49.8    & 62.59  & 74.88  \\
& $\times$    & $\times$      & $\times$       &  86.2    & 80.8    & \textbf{62.85}  & \textbf{74.93}  
\\
\midrule
\multirow{4}{*}{FERV39k} 
& \checkmark   & $\times$     & $\times$       &  51.2    & 42.7    & 40.94  & 50.88     \\
& \checkmark   & \checkmark   & $\times$       &  68.0    & 42.8    & 42.15  & 52.04     \\
& \checkmark   & $\times$     & \checkmark     &  68.1    & 49.6    & 42.25  & 52.01     \\
& \checkmark   & \checkmark   & \checkmark     &  84.9    & 49.8    & 43.12  & 52.07     \\
& $\times$    & $\times$      & $\times$       &  86.2    & 80.8    & \textbf{43.72}  & \textbf{52.52}
\\
\bottomrule
\end{tabular}
}
\end{table}

\begin{figure}[t]
	\centering
    \includegraphics[width=1.0\linewidth]
    {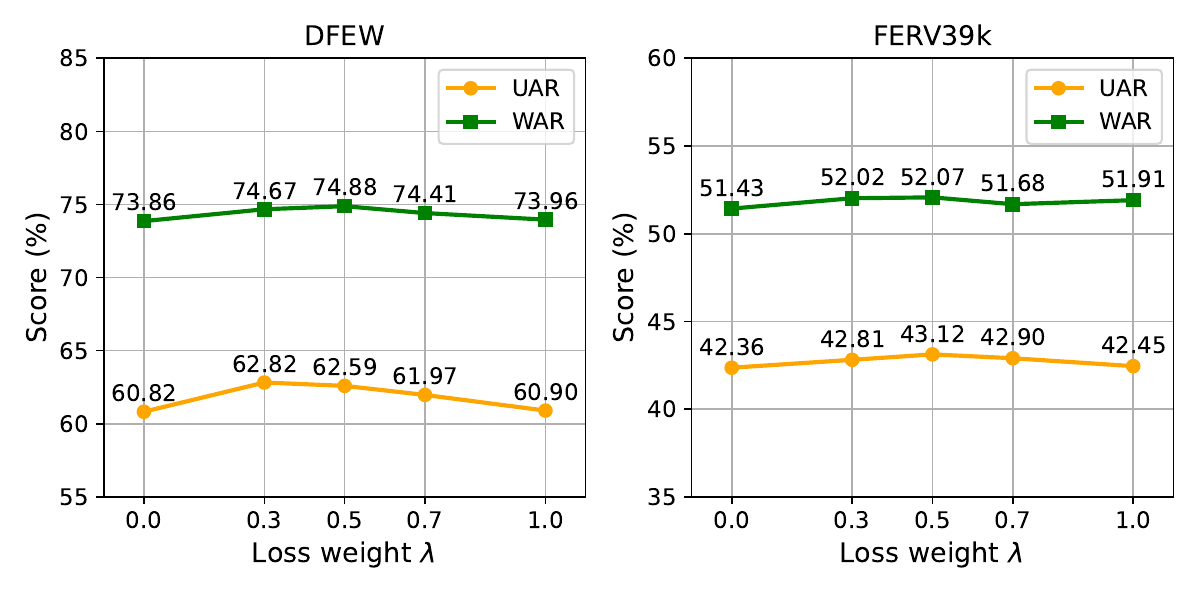}
    \caption{Ablation study on the loss weight.}
	\label{fig_ablation_loss_weight}
\end{figure}

\textbf{Effectiveness of Joint Masked Appearance and Motion Modeling.}
We study the effect of different loss weights in Equation \ref{eq_mae_dfer_loss}, ranging from 1.0 (i.e., only the original appearance target) to 0.0 (i.e., only the motion target). As shown in Fig. \ref{fig_ablation_loss_weight}, we find that the joint model outperforms the model with only one reconstruction target and it achieves the best performance when adopting a loss weight around 0.5. For instance, on DFEW fd1, the best joint model surpasses the standalone appearance model by 1.69\% UAR and 0.92\% WAR and its motion counterpart by 1.77\% UAR and 1.02\% WAR. These results indicate that joint masked appearance and motion modeling are indispensable to facilitate better spatiotemporal representation learning for DFER. 
In addition to our MAE-DFER, we apply it to VideoMAE (shown in Table \ref{tab_ablation_videomae} of Appendix), which can also bring further improvement (1.51\% UAR with 0.30\% WAR on DFEW fd1 and 0.39\% UAR with 0.13\% WAR on FERV39k).

\textbf{Role of Local Region Size.}
We evaluate the effect of different local region sizes in LGI-Former and report the results in Table \ref{tab_ablation_region_size}. We can find that the model performance is not very sensitive to the region size. Moreover, the model computation with different region sizes are similar to each other. These results indicate that, no matter how to divide the input into local regions, LGI-Former can achieve effective and efficient local-global information exchange via the introduced representative tokens and its specialized designs (i.e., the three key modules). Besides, when using the region size of $2\times5\times10$ (only using $M=8$ representative tokens), the model achieves the best performance-computation trade-off.


\begin{table}[]
\caption{Ablation study on the local region size.}
\label{tab_ablation_region_size}
\centering
\resizebox{\linewidth}{!}{
\begin{tabular}{lccccccc}
\toprule
Dataset &  \tabincell{c}{Region size\\($t\times h \times w$)} & $M$ & \tabincell{c}{\#Params (M)}  & \tabincell{c}{FLOPs (G)}   & UAR    & WAR  \\
\midrule
\multirow{4}{*}{DFEW} 
& $1\times5\times10$     & 16 &  84.9    & 49.8    & 62.36  & 74.33     \\
& $2\times2\times10$     & 20 &  84.9    & 50.0    & 61.07  & 74.87     \\
& $2\times5\times10$     &  8 &  84.9    & 49.8    & \textbf{62.59}  & \textbf{74.88}     \\
& $2\times10\times10$    &  4 &  84.9    & 50.7    & 61.27  & 74.19     \\ 
& $4\times5\times10$     &  4 &  84.9    & 50.7    & 62.36  & 74.67     \\
\midrule
\multirow{4}{*}{FERV39k} 
& $1\times5\times10$     & 16 &  84.9    & 49.8    & 42.71  & 52.26     \\
& $2\times2\times10$     & 20 &  84.9    & 50.0    & 42.24  & 52.25     \\
& $2\times5\times10$     &  8 &  84.9    & 49.8    & \textbf{43.12}  & 52.07     \\
& $2\times10\times10$    &  4 &  84.9    & 50.7    & 42.71  & 52.02     \\
& $4\times5\times10$     &  4 &  84.9    & 50.7    & 43.09  & \textbf{52.41}     \\
\bottomrule
\end{tabular}
}
\end{table}



\begin{table}[]
\caption{Results on DFEW. $^\dag$: pre-trained on VoxCeleb2. \underline{Underlined}: the best supervised result. \textit{Bold}: the best result.}
\label{tab_dfew_sota}
\centering
\resizebox{\linewidth}{!}{
\begin{tabular}{lccccccc}
\toprule
Method  &  \tabincell{c}{\#Params (M)} & \tabincell{c}{FLOPs (G)} &  UAR             & WAR \\
\midrule
\multicolumn{4}{l}{\textit{\textbf{Supervised methods}}}   \\
C3D \cite{tran2015learning}             &   78   &  39  &  42.74 & 53.54  \\
R(2+1)D-18 \cite{tran2018closer}        &   33   &  42  &  42.79 & 53.22 \\
3D ResNet-18 \cite{hara2018can}                 &  33  &  8  & 46.52 & 58.27    \\ 
EC-STFL \cite{jiang2020dfew}                    &  -  &  8  & 45.35 & 56.51    \\ 
ResNet-18+LSTM \cite{zhao2021former}              &  -  &  8  & 51.32 & 63.85    \\ 
ResNet-18+GRU  \cite{zhao2021former}           &  -  &  8  & 51.68 & 64.02    \\ 
Former-DFER \cite{zhao2021former}               &  18 &  9  & 53.69 & 65.70   \\ 
CEFLNet \cite{liu2022clip}                     &  13 & -   & 51.14 & 65.35   \\ 
EST \cite{liu2023expression}                    &  43 &  -  & 53.43 & 65.85   \\
STT \cite{ma2022spatio}                           &  -  &  -  & 54.58 & 66.65   \\ 
NR-DFERNet \cite{li2022nr}                     &  -  &  6  & 54.21 & 68.19   \\
DPCNet \cite{wang2022dpcnet}                  &  51 & 10  & \underline{57.11} & 66.32   \\ 
IAL \cite{li2023intensity}               &  19  &  10  & 55.71 & 69.24   \\ 
M3DFEL \cite{wang2023rethinking}         &  -   &  2  & 56.10 & \underline{69.25}       \\
\midrule
\multicolumn{4}{l}{\textit{\textbf{Self-supervised methods}}}   \\
VideoMAE \cite{tong2022videomae}            & 86   &  81 & 58.49  & 70.61 \\ 
VideoMAE  \cite{tong2022videomae} $^\dag$   & 86   &  81 & \textbf{63.60}  & \textbf{74.60} \\ 
MAE-DFER (ours)                             &  85 &  50 & 63.41 & 74.43 \\ 
\bottomrule
\end{tabular}
}
\end{table}

\begin{table}[]
\caption{Results on FERV39k. $^\dag$: pre-trained on VoxCeleb2. \underline{Underlined}: the best supervised result. \textit{Bold}: the best result.}
\label{tab_ferv39k_sota}
\centering
\resizebox{\linewidth}{!}{
\begin{tabular}{lccccccccccc}
\toprule
Method  &  \tabincell{c}{\#Params (M)} & \tabincell{c}{FLOPs (G)} &  UAR             & WAR \\
\midrule
\multicolumn{4}{l}{\textit{\textbf{Supervised methods}}}   \\
C3D \cite{tran2015learning}          &  78  &  39      & 22.68                       & 31.69                      \\
P3D \cite{qiu2017learning}           &  -   &  -      & 30.48                       & 40.81                      \\
R(2+1)D \cite{tran2018closer}        &  -   &  -       & 31.55                       & 41.28                      \\
3D ResNet-18 \cite{hara2018can}       &  33  &  8       & 26.67                       & 37.57                      \\
ResNet-18+LSTM \cite{wang2022ferv39k} &  -   &  -       & 30.92                       & 42.59                      \\
VGG-13+LSTM \cite{wang2022ferv39k}    &  -   &  -       & 32.42                       & 43.37                      \\
Two C3D \cite{wang2022ferv39k}        &  -   &  -       & 30.72                       & 41.77                      \\
Two ResNet-18+LSTM \cite{wang2022ferv39k}     &  -  &  -       & 31.28                       & 43.20                      \\
Two VGG-13+LSTM \cite{wang2022ferv39k}  &  -   &  -      & 32.79                       & 44.54                      \\
Former-DFER \cite{zhao2021former}    &  18  &  9    & 37.20                       & 46.85                 \\
STT \cite{ma2022spatio}              &  -   &  -       & \underline{37.76}           & 48.11                      \\
NR-DFERNet \cite{li2022nr}           &  -   &  6      & 33.99                       & 45.97                      \\
IAL \cite{li2023intensity}       &  19   &  10        & 35.82                       & \underline{48.54}          \\
M3DFEL \cite{wang2023rethinking}         &  -   &  2  & 35.94 & 47.67       \\
\midrule
\multicolumn{4}{l}{\textit{\textbf{Self-supervised methods}}}   \\
VideoMAE \cite{tong2022videomae}           &  86  &  81      & 38.50    & 49.61             \\ 
VideoMAE \cite{tong2022videomae} $^\dag$   &  86  &  81      & \textbf{43.33}   & \textbf{52.39}             \\ 
MAE-DFER (ours)                            &  85  &  50      & 43.12     & 52.07             \\
\bottomrule
\end{tabular}
}
\end{table}

\begin{table}[]
\caption{Results on MAFW. $^\dag$: pre-trained on VoxCeleb2. \underline{Underlined}: the best supervised result. \textit{Bold}: the best result.}
\label{tab_mafw_sota}
\centering
\resizebox{\linewidth}{!}{
\begin{tabular}{lccccccccccccccc}
\toprule
Method & \tabincell{c}{\#Params (M)}  & \tabincell{c}{FLOPs (G)}            & UAR             & WAR            \\ \midrule
\multicolumn{4}{l}{\textit{\textbf{Supervised methods}}}   \\
ResNet-18 \cite{he2016deep}        &  11  &  -   & 25.58 & 36.65          \\
ViT \cite{dosovitskiy2020image}   &  -  &  -   & 32.36 & 45.04          \\
C3D \cite{tran2015learning}       &  78 &  39   & 31.17 & 42.25          \\
ResNet-18+LSTM \cite{liu2022mafw} &  -  &  -   & 28.08           & 39.38          \\
ViT+LSTM \cite{liu2022mafw}       &  -  &  -   & 32.67           & 45.56          \\
C3D+LSTM \cite{liu2022mafw}       &  -  &  -   & 29.75           & 43.76          \\
Former-DFER \cite{zhao2021former}    &  18  &  9    & 31.16	     & 43.27                 \\
T-ESFL \cite{liu2022mafw}        &  -  &  -   & \underline{33.28} & \underline{48.18}     \\
\midrule
\multicolumn{4}{l}{\textit{\textbf{Self-supervised methods}}}   \\
VideoMAE \cite{tong2022videomae}         &  86 &  81 & 38.43 & 51.74        \\
VideoMAE \cite{tong2022videomae} $^\dag$ &  86 &  81 & 40.87 & 53.51        \\
MAE-DFER (ours)                          &  85 &  50 & \textbf{41.62} & \textbf{54.31}        \\
\bottomrule
\end{tabular}
}
\end{table}

\begin{table*}[]
\caption{Results on three lab-controlled datasets. \underline{Underlined}: the best supervised result. \textit{Bold}: the best result.}
\label{tab_lab_controlled_sota}
\centering
\resizebox{0.94\linewidth}{!}{
\begin{tabular}{lccclccclccc}
\toprule
\multicolumn{4}{c}{CREMA-D}  & \multicolumn{4}{c}{RAVDESS} & \multicolumn{3}{c}{eNTERFACE05}\\ 
\cmidrule(lr){1-4} \cmidrule(lr){5-8} \cmidrule(lr){9-11}
Method           &   Modality         & UAR             & WAR            
& Method           &   Modality         & UAR             & WAR            
& Method           & UAR             & WAR            
\\ 
\cmidrule(lr){1-4} \cmidrule(lr){5-8} \cmidrule(lr){9-11}
VO-LSTM \cite{ghaleb2019multimodal}           &   Video          & -                       & \underline{66.80}
& VO-LSTM \cite{ghaleb2019multimodal}           &   Video          & -                       & 60.50
& 3DCNN \cite{byeon2014facial}                                 & -                       & 41.05
\\
Goncalves et al. \cite{goncalves2022robust}   &   Video          & -                       & 62.20  
& 3D ResNeXt-50 \cite{su2020msaf}                &   Video          & -                       & \underline{62.99}
& 3DCNN-DAP \cite{byeon2014facial}                             & -                       & 41.36
\\
Lei et al. \cite{lei2023audio}                &   Video          & \underline{64.68}                   & 64.76  
& AV-LSTM \cite{ghaleb2019multimodal}           &   Video+Audio    & -                       & 65.80
& STA-FER \cite{pan2019adeep}                                  & -                       & 42.98
\\
AV-LSTM \cite{ghaleb2019multimodal}           &   Video+Audio    & -                       & 72.90
& AV-Gating \cite{ghaleb2019multimodal}         &   Video+Audio    & -                       & 67.70
& TSA-FER \cite{pan2019deep}                                   & -                       & 43.72
\\  
AV-Gating \cite{ghaleb2019multimodal}         &   Video+Audio    & -                       & 74.00 
& MCBP \cite{su2020msaf}                        &   Video+Audio    & -                       & 71.32
& C-LSTM \cite{miyoshi2019facial}                              & -                       & 45.29
\\
MulT Base \cite{tran2022pre}                  &   Video+Audio    & -                       & 68.87
& MMTM \cite{su2020msaf}                        &   Video+Audio    & -                       & 73.12
& EC-LSTM \cite{miyoshi2021enhanced}                           & -                       & 49.26
\\ 
MulT Large \cite{tran2022pre}                 &   Video+Audio    & -                       & 70.22
& MSAF \cite{su2020msaf}                        &   Video+Audio    & -                       & 74.86
& FAN \cite{meng2019frame}                                     & -                       & 51.44
\\ 
Goncalves et al. \cite{goncalves2022robust}   &   Video+Audio    & -                       & 77.30
& CFN-SR \cite{fu2021cross}                     &   Video+Audio    & -                       & 75.76
& Graph-Tran \cite{zhao2022spatial}                            & -                       & \underline{54.62}
\\
\cmidrule(lr){1-4} \cmidrule(lr){5-8} \cmidrule(lr){9-11}
MAE-DFER (ours)                                &   Video          & \textbf{77.33}          & \textbf{77.38}  
& MAE-DFER (ours)                                &   Video          & \textbf{75.91}                   & \textbf{75.56}
& MAE-DFER (ours)                                               & \textbf{61.67}                   & \textbf{61.64} 
\\ 
\bottomrule
\end{tabular}
}
\end{table*}


\begin{figure}[t]
	\centering
    \includegraphics[width=1.0\linewidth]{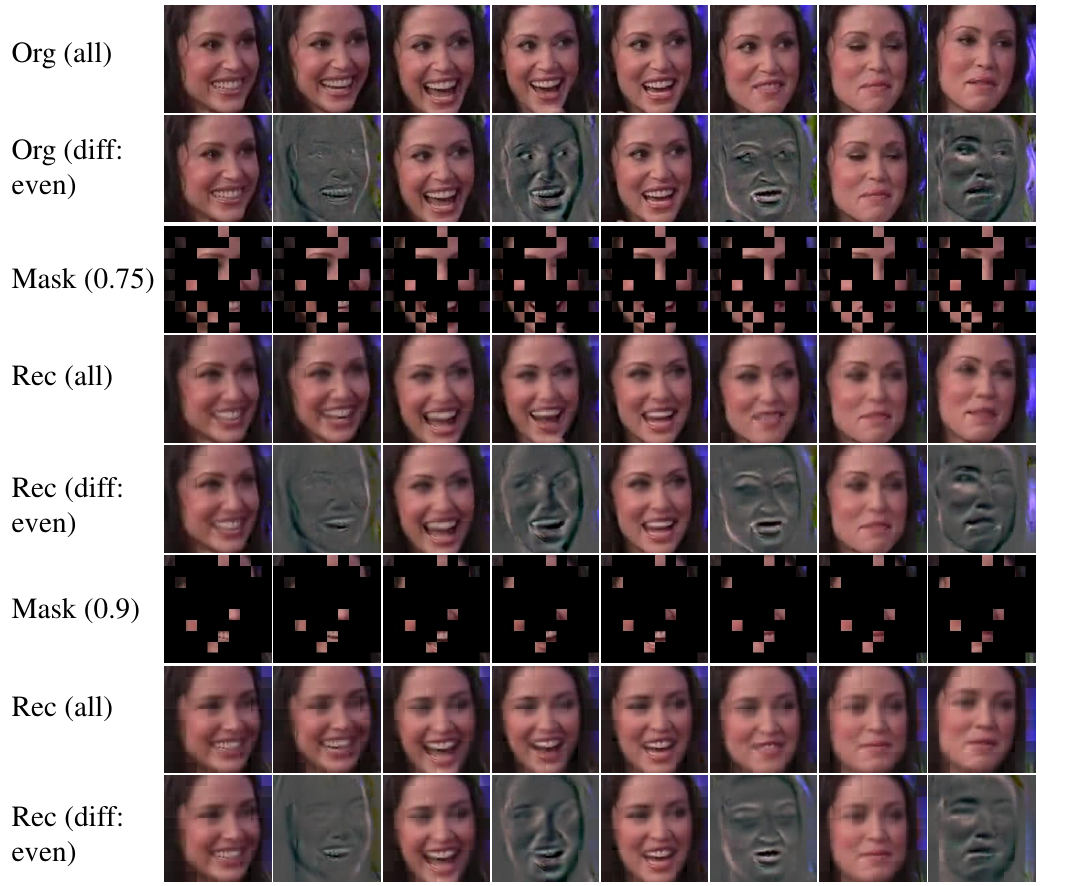}
    \caption{Reconstruction results of a VoxCeleb2 \textit{test} video under masking ratios of 0.75 and 0.9. We only show 8 frames due to the space limitation.
}
	\label{fig_vis_recon}
\end{figure}

\subsection{Comparison with State-of-the-art Methods}

\textbf{Results on Large In-the-wild Datasets.}
We first compare MAE-DFER with previous state-of-the-art supervised methods on DFEW, FERV39k, and MAFW in Table \ref{tab_dfew_sota}, Table \ref{tab_ferv39k_sota}, and Table \ref{tab_mafw_sota}, respectively. 
On DFEW, MAE-DFER surpasses the previous best methods (i.e., DPC-Net \cite{wang2022dpcnet} and M3DFEL \cite{wang2023rethinking}) with a significant margin, achieving a noteworthy \textbf{6.30\%} UAR and \textbf{5.18\%} WAR improvement.
Besides, we also present fine-grained performance of each class in Table \ref{tab_dfew_sota_full_version} of Appendix, MAE-DFER also achieves remarkable improvement across most facial expressions. Notably, for the \textit{disgust} expression, which only accounts for 1.2\% of the entire dataset and is very challenging for all baselines, MAE-DFER improves the best accuracy by over \textbf{10\%}. This considerable improvement indicates that our method is capable of learning powerful representations for DFER via large-scale self-supervised pre-training.
As for the other two datasets, we have similar observations. On the current largest DFER dataset, FERV39k, MAE-DFER achieves the new state-of-the-art performance, exceeding the previous best methods (i.e., STT \cite{ma2022spatio} and IAL \cite{li2023intensity}) by \textbf{5.36\%} UAR and \textbf{3.53\%} WAR. On MAFW, MAE-DFER outperforms the best-performing T-ESFL \cite{liu2022mafw} by a considerable margin of \textbf{8.34\%} UAR and \textbf{6.13\%} WAR. 
Besides, large performance improvement for several rare expressions are also observed on FERV39k and MAFW in Table \ref{tab_ferv39k_sota_full_version} and Table \ref{tab_mafw_sota_full_version} of Appendix. 
In summary, the promising results on three in-the-wild datasets demonstrate the strong generalization ability of MAE-DFER in practical scenarios.

\textbf{Comparison with VideoMAE.}
To verify the effectiveness and efficiency of MAE-DFER, we also show the results of VideoMAE \cite{tong2022videomae} on three in-the-wild datasets, including both the original model pre-trained on Kinetics-400 \cite{carreira2017quo} for 1600 epochs and the model pre-trained on VoxCeleb2 under the same setting as MAE-DFER. 
From Table \ref{tab_dfew_sota}-\ref{tab_mafw_sota}, we have the following observations: 1) The original VideoMAE model pre-trained on \textit{general} videos (i.e., action recognition) is largely inferior to its counterpart pre-trained on \textit{facial} videos, indicating that the large-domain gap between self-supervised pre-training and downstream fine-tuning will severely hurt the performance. 2) Compared with VideoMAE pre-trained on VoxCeleb2, our MAE-DFER largely reduces the computational cost ($\sim$38\% FLOPs) during fine-tuning, while achieving comparable performance on DFEW and FERV39k (only -0.11\%$\sim$0.19\% UAR and -0.17\%$\sim$0.32\% WAR), and even better performance on MAFW (+0.75\% UAR and +0.80\% WAR). Thus, these results demonstrate the effectiveness and efficiency of the proposed method.

\textbf{Results on Small Lab-controlled Datasets.}
We show the comparison results on CREMA-D, RAVDESS, and eNTERFACE05 in Table \ref{tab_lab_controlled_sota}. 
Compared with in-the-wild datasets, we observe \textit{even larger} performance improvement on three lab-controlled datasets. On CREMA-D, our MAE-DFER outperforms the best unimodal methods by over \textbf{12\%} UAR and \textbf{10\%} WAR. More surprisingly, it also shows slightly better performance than the state-of-the-art multimodal method, thus amply demonstrating the superiority of MAE-DFER. On RAVDESS, MAE-DFER improves the previous best by more than \textbf{12\%} WAR and also achieves comparable performance with the best audio-visual method. Finally, on eNTERFACE05, MAE-DFER surpasses the best-performing Graph-Tran \cite{zhao2022spatial} by about \textbf{7\%} WAR.

\begin{figure}[t]
	\centering
    \includegraphics[width=0.9\linewidth]{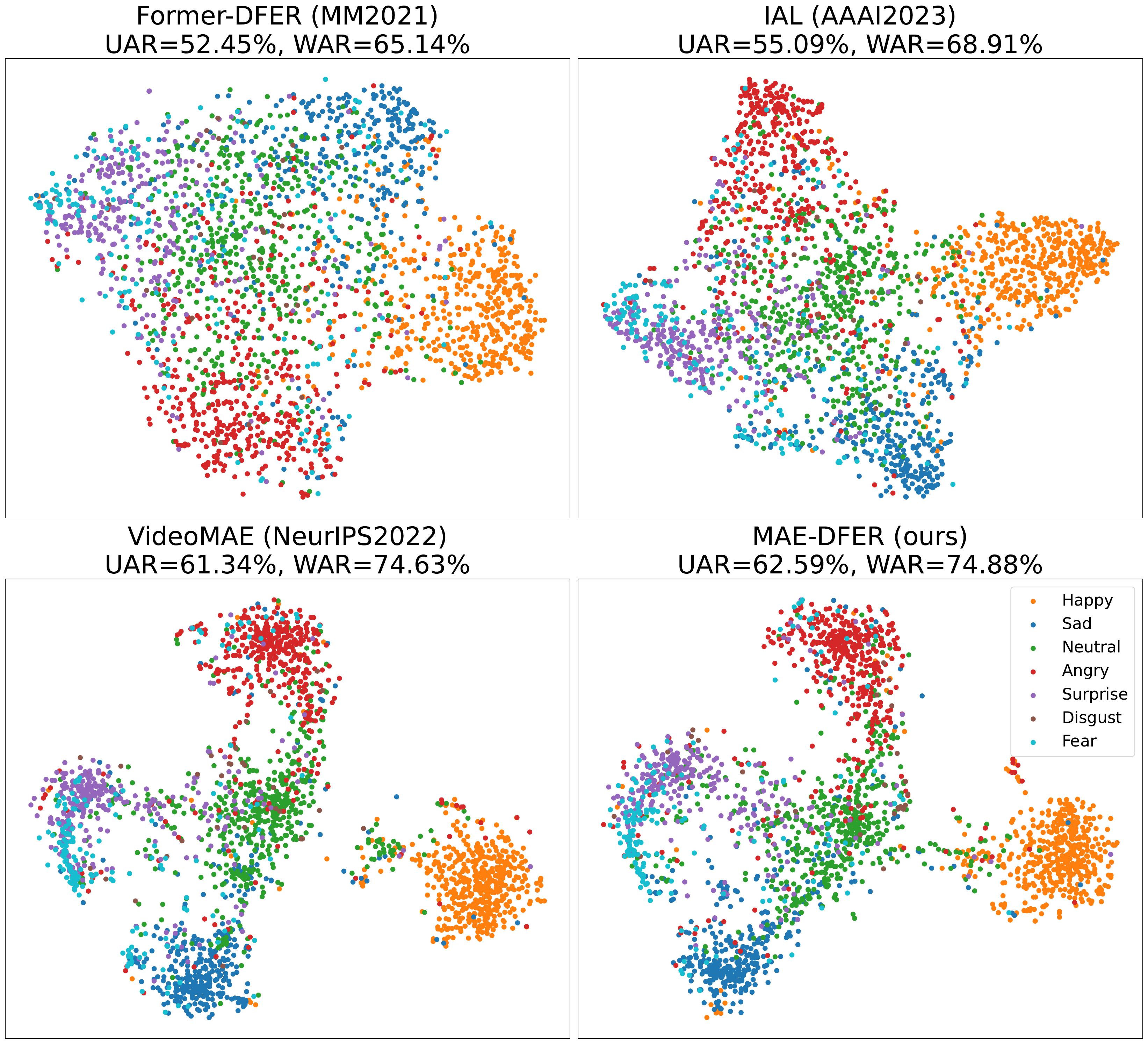}
    \caption{Embedding space visualization using t-SNE \cite{van2008visualizing}.}
	\label{fig_vis_feature}
\end{figure}

\subsection{Visualization Analysis}

\textbf{Reconstruction.}
We first visualize the reconstructed results of MAE-DFER in Fig. \ref{fig_vis_recon}. The video is randomly selected from the VoxCeleb2 \textit{test} set. 
For better visualization, we use a gray-style background for frame difference images shown in \textit{even} frames and also show \textit{all} the reconstructed video by adding the reconstructed frame difference images in \textit{even} frames with the adjacent recovered \textit{odd} frame images. 
From Fig. \ref{fig_vis_recon}, we see that under such a high masking ratio (75\% or 90\%), MAE-DFER still can generate satisfactory reconstructed results for both the facial appearance content and temporal motion information. 
Notably, despite the change in identity information (as the model does not see this person during pre-training), the dynamic facial expression can be well restored by reasoning in limited visible contexts (e.g., the opening mouth).
This imply that our model is able to learn meaningful dynamic facial representations that capture the global spatiotemporal structure.

\textbf{Embedding Space.}
To further qualitatively show the superiority of MAE-DFER over traditional supervised methods, we visualize the learned embeddings using t-SNE \cite{van2008visualizing} on DFEW fd1. As can be seen in Fig. \ref{fig_vis_feature}, the embeddings of our method are more compact and separable than those of two state-of-the-art supervised methods (i.e., IAL \cite{li2023intensity} and Former-DFER \cite{zhao2021former}), which demonstrates that MAE-DFER can learn more discriminative representations for different dynamic facial expressions through large-scale self-supervised pre-training. Besides, VideoMAE has similar embedding space with our MAE-DFER but at the cost of much larger computational cost. 

\section{Conclusion}

In this paper, we have presented an effective and efficient self-supervised framework, namely MAE-DFER, to exploit large amounts of unlabeled facial videos to address the dilemma of current supervised methods and promote the development of DFER. 
We believe MAE-DFER will serve as a strong baseline and foster relevant research in DFER. 
In the future, we plan to explore the scaling behavior of MAE-DFER (i.e., using more data and larger models). Beside, it is also interesting to apply it to other related tasks (e.g., dynamic micro-expression recognition and facial action unit detection).

\begin{acks}
This work is supported by the National Natural Science Foundation of China (NSFC) (No.61831022, No.62276259, No.62201572, No.U21B2010), Beijing Municipal Science \& Technology Commission, Administrative Commission of Zhongguancun Science Park (No.Z211100004821013), Open Research Projects of Zhejiang Lab (No.2021KH0AB06), and CCF-Baidu Open Fund (No.OF2022025).
\end{acks}

\bibliographystyle{ACM-Reference-Format}
\balance
\bibliography{sample-sigconf}

\appendix

\newpage

In Appendix, we provide more information about six DFER dataset, additional ablation studies, and fine-grained results on three in-the-wild datasets.


\section{Datasets}
\label{sec_appendix_dataset}
\textbf{DFEW} comprises 16,372 video clips extracted from over 1,500 high-definition movies. Each video clip is annotated with seven basic emotions (i.e., happy, sad, neutral, anger, surprise, disgust, and fear). We only use 11,697 single-labeled clips in this paper.

\textbf{FERV39k} is currently the largest real-world dynamic facial expression dataset. It has 38,935 video clips with an average length of 1.5 seconds and is annotated with seven basic emotions. 

\textbf{MAFW} is a multimodal compound in-the-wild affective dataset, consisting of 10,045 video clips annotated with 11 compound emotions (including contempt, anxiety, helplessness, disappointment, and seven basic emotions). In this paper, we conduct experiments on 9,172 single-labeled video clips.

\textbf{CREMA-D} is a high-quality audio-visual dataset with 7,442 video clips. Each of them is labeled with six emotions, including happy, sad, anger, fear, disgust, and neutral. 

\textbf{RAVDESS} is an audio-visual dataset that includes emotional speech and song. It consists of  2,880 video clips, each labeled with 8 emotions (i.e., seven basic emotions and calm). In this paper, we only use the speech part consisting of 1,440 video clips. 

\textbf{eNTERFACE05} is an audio-visual emotion recognition dataset that contains approximately 1,200 video clips, each simulating six emotions, including anger, disgust, fear, happy, sad, and surprise. 

\section{More Ablation Studies}
\textbf{Model Size.} 
We investigate the effect of different sizes of LGI-Former to downstream performance. In addition to the default \textit{base} version (512-dim), we also design two smaller versions, i.e., \textit{small} (384-dim) and \textit{tiny} (256-dim). The \textit{small} version has roughly half parameters and FLOPs of the \textit{base} version and it is similar for \textit{tiny} and \textit{small}. As shown in Table \ref{tab_ablation_model_size}, we find that the performance only degrades moderately when the model size becomes smaller, especially for FERV39k. It is worth noting that even the \textit{tiny} version still largely outperforms the state-of-the-art supervised methods (such as DPCNet \cite{wang2022dpcnet} and IAL \cite{li2023intensity} in Table \ref{tab_dfew_sota} and Table \ref{tab_ferv39k_sota}), despite that they has similar parameters and computational cost, which thus further demonstrates the superiority of our proposed method.

\begin{table}[h]
\caption{Ablation study on the model size.}
\label{tab_ablation_model_size}
\centering
\resizebox{\linewidth}{!}{
\begin{tabular}{lccccccc}
\toprule
Dataset &  Size  & Dim & \tabincell{c}{\#Params (M)}  & \tabincell{c}{FLOPs (G)}   & UAR    & WAR  \\
\midrule
\multirow{3}{*}{DFEW} 
& Tiny   & 256   &  21.5    & 13.0    & 59.90  & 73.30     \\
& Small  & 384  &  47.9    & 28.4    & 61.09  & 74.03     \\
& Base   & 512  &  84.9    & 49.8    & \textbf{62.59}  & \textbf{74.88}     \\
\midrule
\multirow{3}{*}{FERV39k} 
& Tiny   & 256   &  21.5    & 13.0    & 41.20  & 51.55     \\
& Small  & 384   &  47.9    & 28.4    & 42.04  & \textbf{52.24}     \\
& Base   & 512   &  84.9    & 49.8    & \textbf{43.12}  & 52.07     \\
\bottomrule
\end{tabular}
}
\end{table}

\textbf{VideoMAE with Joint Masked Appearance and Motion Modeling.} 
Besides our MAE-DFER, we further introduce explicit temporal facial motion modeling to VideoMAE. The results are presented in Table \ref{tab_ablation_videomae}. Similar to our MAE-DFER, we observe that joint masked appearance and motion modeling can further boost the performance of VideoMAE, although standalone motion modeling performs slightly worse than standalone appearance modeling in the original VideoMAE. 

\begin{table}[h]
\caption{Ablation study on VideoMAE with additional temporal facial motion modeling. 
}
\label{tab_ablation_videomae}
\centering
\resizebox{\linewidth}{!}{
\begin{tabular}{lccccccc}
\toprule
Dataset &  Appearance & Motion & \tabincell{c}{\#Params\\(M)}  & \tabincell{c}{FLOPs\\(G)}  & UAR    & WAR  \\
\midrule
\multirow{2}{*}{DFEW} 
& $\times$    & \checkmark  &  86.2  &  80.8 & 60.86  & 74.02     \\
& \checkmark  & $\times$    &  86.2  &  80.8 & 61.34  & 74.63     \\
& \checkmark  & \checkmark  &  86.2  &  80.8 & \textbf{62.85}  & \textbf{74.93}     \\
\midrule
\multirow{2}{*}{FERV39k} 
& $\times$    & \checkmark  &  86.2  &  80.8 & 42.17  & 51.96   \\
& \checkmark  & $\times$    &  86.2  &  80.8 & 43.33  & 52.39   \\
& \checkmark  & \checkmark  &  86.2  &  80.8 & \textbf{43.72}  & \textbf{52.52}     \\
\bottomrule
\end{tabular}
}
\end{table}

\textbf{Role of Classification Token Type.}
We finally explore the effect of two different classification tokens (i.e., original tokens and representative tokens) for downstream fine-tuning. As shown in Table \ref{tab_ablation_cls_token}, we find that performing mean pooling on the representative tokens for final classification slightly outperforms that on the original tokens. We speculate that this is because the representative tokens are more compact and high-level than the original tokens.

\begin{table}[h]
\caption{Ablation study on the classification token type.}
\label{tab_ablation_cls_token}
\centering
\scalebox{0.9}{
\begin{tabular}{ccccccc}
\toprule
\multirow{2}{*}{Token type}  & \multicolumn{2}{c}{DFEW} & \multicolumn{2}{c}{FERV39k} \\
          \cmidrule(lr){2-3}     \cmidrule(lr){4-5}
        & UAR    & WAR    & UAR    & WAR  \\
\midrule
Original tokens         & 62.16  & 74.51  & 42.89  & 51.91   \\
Representative tokens   & \textbf{62.59}  & \textbf{74.88}  & \textbf{43.12}  & \textbf{52.07}   \\
\bottomrule
\end{tabular}
}
\end{table}

\section{Detailed Results}
In this section, we first present more fine-grained results (i.e., accuracy of each class) on DFEW, FERV39k, and MAFW in Table \ref{tab_dfew_sota_full_version}, Table \ref{tab_ferv39k_sota_full_version}, and Table \ref{tab_mafw_sota_full_version}, respectively. 
From three tables, we observe that MAE-DFER significantly outperforms the state-of-the-art supervised methods on most facial expressions, especially on some \textit{rare} facial expressions (such as \textit{disgust}, \textit{contempt}, and \textit{disappointment}). 
For instance, on DFEW, our MAE-DFER surpasses the previous best supervised results by about 9\% on \textit{sad}, 13\% on \textit{disgust}, and 8\% on \textit{fear}. 
On MAFW, it improves the best-performing supervised methods by over 5\% on \textit{anger}, 7\% on \textit{disgust}, 8\% on \textit{contempt}, 8\% on \textit{anxiety},  6\% on \textit{helplessness}, and 7\% on \textit{disappointment}.
Moreover, compared with VideoMAE pre-trained under the same setting, MAE-DFER has comparable or even better fine-grained performance while largely reduces the computational cost during fine-tuning. 
We also note that the original VideoMAE pre-trained on Kinetics-400 does not perform well on some rare expressions (e.g., \textit{disgust} on FERV39k), although it could achieve the best results on some dominated expressions (e.g., \textit{neutral} on FERV39k).
These results indicate that our MAE-DFER can effectively and efficiently learn more robust and general representations for DFER via large-scale self-supervised training on abundant unlabeled facial videos, thus mitigating the unbalanced learning issue and achieving superior fine-grained performance. 


\begin{table*}[]
\caption{Results on DFEW. $^\dag$: pre-trained on VoxCeleb2. \underline{Underlined}: the best supervised result. \textit{Bold}: the best result.
}
\label{tab_dfew_sota_full_version}
\centering
\scalebox{0.8}{
\begin{tabular}{lccccccccccc}
\toprule
\multirow{2}{*}{Method} & \multirow{2}{*}{\tabincell{c}{\#Params\\(M)}}  & \multirow{2}{*}{\tabincell{c}{FLOPs\\(G)}}  
& \multicolumn{7}{c}{Accuracy of Each Emotion (\%)}  & \multicolumn{2}{c}{Metric (\%)} \\ \cmidrule(lr){4-10} \cmidrule(lr){11-12} 
&                            &                       & Happy          & Sad            & Neutral        & Anger          & Surprise       & Disgust       & Fear           & UAR             & WAR            \\ \midrule
\multicolumn{12}{l}{\textit{\textbf{Supervised methods}}}   \\
C3D \cite{tran2015learning}             &   78   &  39 & 75.17 & 39.49 & 55.11   & 62.49 & 45.00    & 1.38    & 20.51 &  42.74 & 53.54  \\
R(2+1)D-18 \cite{tran2018closer}        &   33   &  42 & 79.67 & 39.07 & 57.66   & 50.39 & 48.26    & 3.45    & 21.06 &  42.79 & 53.22 \\
3D ResNet-18 \cite{hara2018can}                     &  33 &  8  & 76.32 & 50.21 & 64.18 & 62.85 & 47.52 & 0.00 & 24.56 & 46.52 & 58.27    \\ 
EC-STFL \cite{jiang2020dfew}                        &  -  &  8  & 79.18 & 49.05 & 57.85 & 60.98 & 46.15 & 2.76 & 21.51 & 45.35 & 56.51    \\ 
ResNet-18+LSTM \cite{zhao2021former}                &  -  &  8  & 83.56 & 61.56 & 68.27 & 65.29 & 51.26 & 0.00 & 29.34 & 51.32 & 63.85    \\ 
ResNet-18+GRU  \cite{zhao2021former}                &  -  &  8  & 82.87 & 63.83 & 65.06 & 68.51 & 52.00 & 0.86 & 30.14 & 51.68 & 64.02    \\ 
Former-DFER \cite{zhao2021former}                   &  18 &  9  & 84.05 & 62.57 & 67.52 & 70.03 & 56.43 & 3.45 & 31.78 & 53.69 & 65.70   \\ 
CEFLNet \cite{liu2022clip}                          &  13 & -   & 84.00 & 68.00 & 67.00 & 70.00 & 52.00 & 0.00 & 17.00 & 51.14 & 65.35   \\ 
EST \cite{liu2021expression}                        &  43 &  -  & 86.87 & 66.58 & 67.18 & 71.84 & 47.53 & \underline{5.52} & 28.49 & 53.43 & 65.85   \\
STT \cite{ma2022spatio}                             &  -  &  -  & 87.36 & 67.90 & 64.97 & 71.24 & 53.10 & 3.49 & \underline{34.04} & 54.58 & 66.65   \\ 
NR-DFERNet \cite{li2022nr}                          &  -  &  6  & 88.47 & 64.84 & 70.03 & 75.09 & 61.60 & 0.00 & 19.43 & 54.21 & 68.19   \\ 
DPCNet \cite{wang2022dpcnet}                        &  51 & 10  & -     & -     & -     & -     & -     & -    & -     & \underline{57.11} & 66.32   \\ 
IAL \cite{li2023intensity}                      &  19  &  10  & 87.95 & 67.21 & \underline{70.10} & \underline{76.06} & \underline{62.22} & 0.00 & 26.44 & 55.71 & 69.24   \\
M3DFEL \cite{wang2023rethinking}         &  -   &  2  & \underline{89.59} & \underline{68.38} & 67.88 & 74.24 & 59.69 & 0.00 & 31.63 & 56.10 & \underline{69.25}       \\
\midrule
\multicolumn{12}{l}{\textit{\textbf{Self-supervised methods}}} \\
VideoMAE \cite{tong2022videomae}           &  86   &  81  & 92.23 & 67.81 & 70.97 & 74.02 & 62.59 & 10.34 & 31.49 & 58.49  & 70.61 \\ 
VideoMAE \cite{tong2022videomae} $^\dag$   &  86   &  81  & \textbf{93.09} & \textbf{78.78} & 71.75 & \textbf{78.74} & \textbf{63.44} & 17.93 & 41.46 & \textbf{63.60}  & \textbf{74.60} \\ 
MAE-DFER (ours)                            &  85   &  50  & 92.92 & 77.46 & \textbf{74.56} & 76.94 & 60.99 & \textbf{18.62} & \textbf{42.35} & 63.41  & 74.43 \\ 
\bottomrule
\end{tabular}
}
\end{table*}

\begin{table*}[]
\caption{Results on FERV39k. $^\dag$: pre-trained on VoxCeleb2.
\underline{Underlined}: the best supervised result. \textit{Bold}: the best result.
}
\label{tab_ferv39k_sota_full_version}
\centering
\scalebox{0.8}{
\begin{tabular}{lccccccccccc}
\toprule
\multirow{2}{*}{Method} & \multirow{2}{*}{\tabincell{c}{\#Params\\(M)}}  & \multirow{2}{*}{\tabincell{c}{FLOPs\\(G)}}  
& \multicolumn{7}{c}{Accuracy of Each Emotion (\%)}  & \multicolumn{2}{c}{Metric (\%)} \\ \cmidrule(lr){4-10} \cmidrule(lr){11-12} 
                        &  &  & Happy          & Sad            & Neutral        & Anger          & Surprise       & Disgust       & Fear           & UAR             & WAR            \\ \midrule
\multicolumn{12}{l}{\textit{\textbf{Supervised methods}}}   \\
C3D \cite{tran2015learning}          &  78  &  39   & 48.20   & 35.53   & 52.71     & 13.72   & 3.45       & 4.93     & 0.23    & 22.68                       & 31.69                      \\
P3D \cite{qiu2017learning}           &  -   &  -   & 61.85   & 42.21   & 49.80     & 42.57   & 10.50      & 0.86     & 5.57    & 30.48                       & 40.81                      \\
R(2+1)D \cite{tran2018closer}        &  -   &  -   & 59.33   & 42.43   & 50.82     & 42.57   & 16.30      & 4.50     & 4.87    & 31.55                       & 41.28                      \\
3D ResNet-18 \cite{hara2018can}       &  33  &  8   & 57.64   & 28.21   & 59.60     & 33.29   & 4.70       & 0.21     & 3.02    & 26.67                       & 37.57                      \\
ResNet-18+LSTM \cite{wang2022ferv39k} &  -   &  -   & 61.91   & 31.95   & 61.70     & 45.93   & 14.26      & 0.00     & 0.70    & 30.92                       & 42.59                      \\
VGG-13+LSTM \cite{wang2022ferv39k}    &  -   &  -   & 66.26   & 51.26   & 53.22     & 37.93   & 13.64      & 0.43     & 4.18    & 32.42                       & 43.37                      \\
Two C3D \cite{wang2022ferv39k}        &  -   &  -   & 54.85   & 52.91   & 60.67     & 31.34   & 5.96       & 2.36     & 6.96    & 30.72                       & 41.77                      \\
Two ResNet-18+LSTM \cite{wang2022ferv39k}     &  -  &  -   & 59.00   & 45.87   & \underline{61.90}     & 40.15   & 9.87       & 1.71     & 0.46    & 31.28                       & 43.20                      \\
Two VGG-13+LSTM \cite{wang2022ferv39k}  &  -   &  -   & 69.65   & 47.31   & 52.55     & 47.88   & 7.68       & 1.93     & 2.55    & 32.79                       & 44.54                      \\
Former-DFER \cite{zhao2021former}    &  18  &  9   & 65.65   & 51.33   & 56.74     & 43.64   & \underline{21.94}      & 8.57     & \underline{12.53}   & 37.20                       & 46.85                 \\
STT \cite{ma2022spatio}              &  -   &  -   & \underline{69.77}   & 47.81   & 59.14     & 47.41   & 20.22      & \underline{10.49}    & 9.51    & \underline{37.76}           & 48.11                      \\
NR-DFERNet \cite{li2022nr}           &  -   &  6   & 69.18   & \underline{\textbf{54.77}}   & 51.12     & \underline{49.70}   & 13.17      & 0.00     & 0.23    & 33.99                       & 45.97                      \\
IAL \cite{li2023intensity}       &  19   &  10   &  -      &  -      &  -        &  -      &  -         &  -       &  -      & 35.82                       & \underline{48.54}          \\ 
M3DFEL \cite{wang2023rethinking}         &  -   &  2  &  -      &  -      &  -        &  -      &  -         &  -       &  -        & 35.94 & 47.67       \\
\midrule
\multicolumn{12}{l}{\textit{\textbf{Self-supervised methods}}}   \\
VideoMAE \cite{tong2022videomae}           &  86  &  81  & 71.28   & 48.60   & \textbf{63.99}     & 47.28   & 20.69      & 5.35    &  12.30        & 38.50    & 49.61             \\ 
VideoMAE \cite{tong2022videomae} $^\dag$   &  86  &  81  & 72.91   & 54.34   & 59.50     & \textbf{51.65}   & 29.47      & 17.77    &  \textbf{17.63}  & \textbf{43.33}                       & \textbf{52.39}             \\ 
MAE-DFER (ours)                            &  85  &  50  & \textbf{73.05}   & 53.98   & 59.14     & 50.44   & \textbf{30.09}      & \textbf{17.99}    &  17.17  & 43.12                       & 52.07             \\
\bottomrule
\end{tabular}
}
\end{table*}

\begin{table*}[]
\caption{Results on MAFW. $^\dag$: pre-trained on VoxCeleb2. \underline{Underlined}: the best supervised result. \textit{Bold}: the best result.
}
\label{tab_mafw_sota_full_version}
\centering
\scalebox{0.8}{
\begin{tabular}{lccccccccccccccc}
\toprule
\multirow{2}{*}{Method} & \multirow{2}{*}{\tabincell{c}{\#Params\\(M)}}  & \multirow{2}{*}{\tabincell{c}{FLOPs\\(G)}}  
& \multicolumn{11}{c}{Accuracy of Each Emotion (\%)}  & \multicolumn{2}{c}{Metric (\%)} \\ 
\cmidrule(lr){4-14} \cmidrule(lr){15-16} 
&  &  & AN    & DI    & FE    & HA    & NE    & SA    & SU    & CO   & AX    & HL   & DS   & UAR  & WAR  \\ 
\midrule
ResNet-18 \cite{he2016deep}        &  11  &  -  & 45.02 & 9.25  & 22.51 & 70.69 & 35.94 & 52.25 & 39.04 & 0.00 & 6.67  & 0.00 & 0.00 & 25.58 & 36.65          \\
ViT \cite{dosovitskiy2020image}   &  -  &  -  & 46.03 & \underline{18.18} & 27.49 & 76.89 & 50.70 & 68.19 & 45.13 & 1.27 & 18.93 & 1.53 & 1.65 & 32.36 & 45.04          \\
C3D \cite{tran2015learning}       &  78 &  39  & 51.47 & 10.66 & 24.66 & 70.64 & 43.81 & 55.04 & 46.61 & 1.68 & 24.34 & \underline{5.73} & \underline{4.93} & 31.17 & 42.25          \\
ResNet-18+LSTM \cite{liu2022mafw} &  -  &  -  & 46.25 & 4.70  & 25.56 & 68.92 & 44.99 & 51.91 & 45.88 & \underline{1.69} & 15.75 & 1.53 & 1.65 & 28.08           & 39.38          \\
ViT+LSTM \cite{liu2022mafw}       &  -  &  -  & 42.42 & 14.58 & \underline{\textbf{35.69}} & 76.25 & 54.48 & \underline{68.87} & 41.01 & 0.00 & 24.40 & 0.00 & 1.65 & 32.67           & 45.56          \\
C3D+LSTM \cite{liu2022mafw}       &  -  &  -  & 54.91 & 0.47  & 9.00  & 73.43 & 41.39 & 64.92 & \underline{\textbf{58.43}} & 0.00 & \underline{24.62} & 0.00 & 0.00 & 29.75           & 43.76          \\
Former-DFER \cite{zhao2021former}    &  18  &  9  &  -      &  -      &  -        &  -      &  -         &  -       &  -   &  -      &  -      &  -        &  -       & 31.16	     & 43.27     \\
T-ESFL \cite{liu2022mafw}        &  -  &  -  & \underline{62.70} & 2.51  & 29.90 & \underline{\textbf{83.82}} & \underline{\textbf{61.16}} & 67.98 & 48.50 & 0.00 & 9.52  & 0.00 & 0.00 & \underline{33.28} & \underline{48.18}     \\
\midrule
\multicolumn{16}{l}{\textit{\textbf{Self-supervised methods}}}   \\
VideoMAE \cite{tong2022videomae}       &  86 &  81 & 62.23   & 23.32 & 32.64 & 78.18 & 60.28 & 66.60 & 56.81 & 0.41 & 27.62 & 5.34 & 8.24 & 38.34 & 51.74        \\
VideoMAE \cite{tong2022videomae} $^\dag$ &  86 &  81 & 65.90 & 23.63 & 34.88 & 76.73 & 55.62 & \textbf{73.47} & 54.57 & \textbf{9.75} & 32.75 & 10.69 & 11.54 & 40.87 & 53.51        \\
MAE-DFER (ours)                          &  85 &  50 & \textbf{67.77} & \textbf{25.35} & 34.88 & 77.13 & 58.26 & 71.09 & 57.46 & 8.90 & \textbf{33.08} & \textbf{11.83} & \textbf{12.09} & \textbf{41.62} & \textbf{54.31}        \\
\bottomrule
\end{tabular}
}
\end{table*}

\end{document}